\documentclass[journal]{IEEEtran}

\usepackage[linesnumbered,lined,boxed,commentsnumbered,ruled]{algorithm2e}
\SetKwComment{Comment}{$\triangleright$ }{}
\SetKwInOut{Input}{Input}\SetKwInOut{Output}{Output}
\usepackage{etoolbox}

\SetCommentSty{mycommfont}

\makeatletter
\patchcmd{\@algocf@start}
  {-1.5em}
  {0pt}
  {}{}
\makeatother

\usepackage{amsmath,amssymb,amsfonts}
\usepackage[usenames,dvipsnames]{xcolor} 
\usepackage{multirow}
\usepackage{float}
\usepackage{pdfpages}
\usepackage{amsmath}
\usepackage{comment}
\usepackage{float}
\usepackage{bbm}
\usepackage{multicol}
\usepackage{adjustbox}
\usepackage{amssymb}
\usepackage{amsfonts}
\usepackage{multicol}
\usepackage{pifont}
\usepackage{makecell}
\def\cmark{\ding{51}} 
\def\xmark{\ding{55}}
\usepackage{mathtools}
\usepackage{upgreek}
\usepackage{url}
\usepackage{pgfplots}
\usepackage[noadjust]{cite}
\usepgfplotslibrary{colorbrewer}
\usepackage{soul}

\usepackage{pgfplots}  
\usepackage{pgfplotstable}  
\usepackage{xcolor}  
\usepackage{graphicx}  
\usetikzlibrary{intersections}  
\usetikzlibrary{patterns}  
\usetikzlibrary{plotmarks}  %
\colorlet{FedProx_c}{Brown}

\DeclareMathAlphabet\mathbfcal{OMS}{cmsy}{b}{n}

\title{A Multi-Modal Federated Learning Framework for Remote Sensing Image Classification}

\begin{document}

\author{Barı\c{s} B\"{u}y\"{u}kta\c{s},
        Gencer Sumbul,~\IEEEmembership{Member,~IEEE},
        Beg\"{u}m Demir,~\IEEEmembership{Senior Member,~IEEE}%
        \thanks{Barı\c{s} B\"{u}y\"{u}kta\c{s} and Beg{\"u}m Demir are with the Faculty of Electrical Engineering and Computer Science, Technische Universit\"at Berlin, 10623 Berlin, Germany, also with the BIFOLD - Berlin Institute for the Foundations of Learning and Data, 10623 Berlin, Germany.
        Email: \mbox{baris.bueyuektas@tu-berlin.de}, \mbox{demir@tu-berlin.de}.
        
        Gencer Sumbul is with the Environmental Computational Science and Earth Observation Laboratory (ECEO), École Polytechnique Fédérale de Lausanne (EPFL), 1950 Sion, Switzerland (e-mail: \mbox{gencer.sumbul@epfl.ch}).}%
}

\markboth{Journal of \LaTeX\ Class Files,~Vol.~14, No.~8, August~2015}%
{Shell \MakeLowercase{\textit{et al.}}: Bare Demo of IEEEtran.cls for IEEE Journals}
\maketitle
\begin{abstract}

Federated learning (FL) enables the collaborative training of deep neural networks across decentralized data archives (i.e., clients) without sharing the local data of the clients. Most of the existing FL methods assume that the data distributed across all clients is associated with the same data modality. However, remote sensing (RS) images present in different clients can be associated with diverse data modalities. The joint use of the multi-modal RS data can significantly enhance classification performance. To effectively exploit decentralized and unshared multi-modal RS data, our paper introduces a novel multi-modal FL framework for RS image classification problems. The proposed framework comprises three modules: 1) multi-modal fusion (MF); 2) feature whitening (FW); and 3) mutual information maximization (MIM). The MF module employs iterative model averaging to facilitate learning without accessing multi-modal training data on clients. The FW module aims to address the limitations of training data heterogeneity by aligning data distributions across clients. The MIM module aims to model mutual information by maximizing the similarity between images from different modalities. For the experimental analyses, we focus our attention on multi-label classification and pixel-based classification tasks in RS. The results obtained using two benchmark archives show the effectiveness of the proposed framework when compared to state-of-the-art algorithms in the literature. The code of the proposed framework will be available at https://git.tu-berlin.de/rsim/multi-modal-FL.

\end{abstract}


\IEEEpeerreviewmaketitle

\section{Introduction}

Due to the advances in satellite technology, there has been a significant increase in the amount of remote sensing (RS) images distributed across decentralized image archives (i.e., clients). To achieve accurate knowledge discovery from RS image archives, deep learning (DL) methods have demonstrated significant success in RS, requiring complete access of training data, while learning the DL model parameters during training \cite{lyu2022privacy}. This requirement may not suit to some operational RS applications, since images in different clients (decentralized data archives) can be unshared due to commercial concerns, legal regulations, and privacy. In RS, commercial data providers often view their data as a valuable asset. In detail, these providers earn income by selling access to their data, and thus they restrict free data access to protect their business interests. There may also be regulatory constraints or licensing agreements that hinders RS data providers from openly disseminating their data to the public. Also, legal constraints, such as privacy laws and national security considerations, could also prevent public access of training data on clients \cite{zhang2022progress}. Privacy is particularly important for certain RS applications such as crop monitoring, damage assessment, and wildlife tracking. For instance, farmers may not be interested to share labels of different crops \cite{tuia2023artificial}. RS images used to evaluate the impact of a disaster \cite{munawar2022remote} may include personal and sensitive information. As an other example, in the context of wildlife tracking through RS images \cite{rafiq2023sensordrop}, training data is often kept confidential to prevent the revelation of precise locations of endangered species or sensitive ecosystems.


To train DL models on unshared image archives, several federated learning (FL) methods, which aim to find the optimal model parameters on a central server (i.e., global model) without accessing data on clients, have been proposed. Most existing FL methods assume that the images on different clients are associated with a single data modality (see Section \ref{lit}). However, the images on different clients can belong to different data modalities. The multi-modal images associated with the same geographical area allow for a rich characterization of RS images when jointly considered, and thus improve the effectiveness of the considered image analysis task \cite{yao2022multi}. To jointly exploit multi-modal RS images, the development of multi-modal image classification methods has attracted great attention in RS \cite{hoffmann2023transformer,fang2023classification,weilandt2023early}. However, these methods assume that all the multi-modal images are accessible during training, and thus they are not suitable to be used in the context of FL. 

The adaptation of existing FL algorithms in RS is not fully suitable to accurately learn from data when the clients are associated with different modalities. This is due to the fact that the different modalities necessitate different architectures to extract the meaningful features \cite{zhang2023meta}. Since the number of parameters kept in the clients is different under different architectures, it is not always possible to take an average during the aggregation phase as in many methods in the literature \cite{moshawrab2023reviewing}. Even if the training could be done using the same architectures on the clients with different modalities, the convergence of global model is impeded after the aggregation. This is because the clients associated with different modalities often have distinct data representations, posing a challenge in deriving a unified central model capable of encapsulating the diverse characteristics present in data on clients. To tackle these challenges, the development of multi-modal (MM) FL methods is needed. This topic is not studied yet in RS,  while widely investigated in computer vision (CV) \cite{yu2023multimodal,xiong2022unified,chen2022fedmsplit} (see Section \ref{lit} for a detailed review). Most of the existing MM FL works in CV are either associated to the significant communication overhead \cite{chen2022fedmsplit} or assume that an auxiliary data exist during training \cite{yu2023multimodal,xiong2022unified}. However, in operational scenarios in RS, the reliable auxiliary data (which accurately represents the data distributed to the clients) may not be available, leading to insufficient FL performance.

To address these issues, in this paper we introduce a novel multi-modal FL framework that learns model parameters from decentralized multi-modal RS image archives without accessing data on clients, while ensuring communication efficiency without any need for leveraging auxiliary data. The proposed framework is made up of three main modules: 1) multi-modal fusion (MF); 2) feature whitening (FW); and 3) mutual information maximization (MIM). The MF module aims to employ FL through iterative model averaging when clients are associated with different data modalities. It is achieved by defining modality-specific backbones and a common classifier on the central server. The FW module aims to overcome the limitations of training data heterogeneity across clients. This is accomplished by using batch whitening layers instead of batch normalization layers. The MIM module aims to capture the mutual information between distinct modalities by maximizing the similarity of images acquired from the same geographical area but associated with different data modalities. This is achieved by maximizing the similarity between the feature vectors extracted from the local model and those from the aggregated models through contrastive learning. While these techniques have been explored in centralized learning settings, their adaptation has not been previously addressed in FL. We extend and modify these modules to function effectively in a decentralized, privacy-preserving FL setting. The MF module introduces a novel FL-specific fusion mechanism that enables learning across unshared multi-modal data without requiring direct feature exchange. The FW module adapts batch whitening to FL, ensuring alignment of feature distributions across non-IID clients while maintaining privacy. The MIM module is uniquely designed for FL by leveraging global model representations to approximate cross-modal alignment without access to paired multi-modal samples. These adaptations enable effective multi-modal learning in FL. It is worth noting that the proposed framework has been briefly introduced in \mbox{\cite{Buyuktas:2023}} with limited experimental results. In this paper, we extend our work by providing a comprehensive description of the proposed framework. Moreover, we conduct extensive experimental analysis in the context of multi-label scene classification \mbox{\cite{burgert2022effects,aksoy2022multi}} using BigEarthNet-MM \mbox{\cite{sumbul2021bigearthnet}} and pixel-based classification \mbox{\cite{bruzzone2014review}} using Dynamic World-Expert \mbox{\cite{fuller2024croma}}. In addition, we compare the proposed framework  with several state-of-the-art FL algorithms. We would like to note that our framework is independent from the considered number of image modalities as well as their associated architectures. To the best of our knowledge, it is the first RS FL framework that learns DL model parameters from decentralized and unshared multi-modal training data in the framework of RS image classification problems.

The rest of this article is organized as follows. Section II presents the related works. Section III introduces the proposed multi-modal FL framework. Section IV describes the considered RS image archives and the experimental setup, while Section V provides the experimental results. Section VI concludes our article.


\section{Literature Review}
\label{lit}

The existing FL algorithms can be grouped into two different categories: 1) single-modal FL algorithms (which assume that each client is associated with the same modality of the data); 2) multi-modal FL algorithms (which consider that clients can be associated with the different modalities of the data). In the following subsections, we review the single-modal and multi-modal FL algorithms existing both in CV and RS communities.

\subsection{Single-modal FL Algorithms}

The development of single-modal FL algorithms has been widely studied in CV. As an example, in~\cite{li2020federated}, a federated optimization algorithm (denoted as FedProx) is proposed to control the local updates by minimizing Euclidean distances between the parameters of the global model and each local model. In~\cite{li2020federated}, a model-contrastive FL algorithm (denoted as MOON) is introduced to improve the generalization capability of the local model by reducing the distance of feature distributions among clients and central model. In \cite{karimireddy2020scaffold}, the stochastic controlled averaging algorithm (denoted as SCAFFOLD) is introduced to minimize the divergence between the update direction of local models and central model by incorporating a correction mechanism that adjusts local updates. In \cite{gao2022feddc}, a FL algorithm with local drift decoupling and correction (denoted as FedDC) is proposed to reduce the difference between the local model and the global model parameters. In \cite{li2021fedbn}, a FL algorithm (denoted as FedBN) is proposed to alleviate the feature shift between clients. To this end, batch normalization parameters are kept at the clients and not sent to the global server for aggregation. The algorithm proposed in \cite{ma2022layer}, known as the layer-wised personalized federated learning algorithm (denoted as pFedLA), assigns an importance score for each layer from different clients and the aggregation is done per client according to the importance scores that are obtained from different clients. In \cite{wu2022communication}, a FL algorithm (denoted as FedKD) is proposed to reduce the number of shared parameters to the central server by applying adaptive mutual knowledge distillation techniques. 

While FL has been extensively studied in CV, it is seldom considered in the field of RS \cite{zhang2023federated, zhang2022prototype, Buyuktas:2023}. In \cite{zhang2023federated}, a FL scheme with prototype matching algorithm (denoted as FedPM) is introduced. This algorithm employs a prototype matching algorithm to address data distribution divergence. It generates prototypical class representations on each client and then aggregates these representations at the central server for pixel-based remote sensing image classification. In \cite{zhang2022prototype}, a FL algorithm combined with prototype-based hierarchical clustering (denoted as Fed-PHC) is introduced for pixel-based RS image classification to tackle data heterogeneity by clustering clients based on data distribution similarities and employing personalized model averaging. In \cite{buyuktacs2023federated}, state-of-the-art FL algorithms that address the limitations of training data heterogeneity are analyzed under different decentralization scenarios in the context of multi-label image classification in RS. In \cite{buyuktacs2024transformer}, the integration of transformer architectures into FL is explored to improve multi-label classification of RS images, particularly addressing non-IID data across clients. In \cite{li2024towards}, a framework is proposed to obtain personalized global models specific to different RS data characteristics and tasks. In \cite{wang2024personalized}, a personalized adaptive distillation (PAD) scheme is introduced to prevent overfitting in few-shot RS classification by adaptively aggregating local model layers. In addition to addressing the limitations of data heterogeneity, recent research in FL for RS has also focused on reducing communication cost and enhancing privacy. In \cite{chen2024free}, a privacy-preserving FL framework (denoted as PRFL) is introduced for fine-grained RS target classification to reduce communication costs via dynamic parameter decomposition. In \cite{zhang2023local}, the local differential privacy embedded FL algorithm (denoted as LDP-Fed) is proposed to apply local differential privacy to protect data and model privacy against membership inference attacks in RS scene classification. In \cite{zhu2023privacy}, a blockchain-based FL algorithm is introduced to protect DL models from poisoning attacks in RS image classification.

\subsection{Multi-modal FL Algorithms}
In recent years, several multi-modal FL algorithms have been developed in CV. As an example, in \cite{yu2023multimodal}, the contrastive representation ensemble and aggregation for multi-modal FL (denoted as CreamFL) algorithm is introduced to learn from decentralized clients that might include image and text data by communicating representations between clients and a central server, utilizing auxiliary data. The knowledge distillation is used for each modality to learn a global model. To achieve a better fusion of multi-modal representations, inter-modal and intra-modal contrastive regularization techniques are used in this algorithm. In \cite{qayyum2022collaborative}, a clustered FL algorithm integrates medical images and clinical data to improve COVID-19 diagnosis performance while leveraging the capabilities of edge computing. In \cite{che2023multimodal}, an algorithm is proposed that assumes the clients include unlabelled data and the central server includes auxiliary labelled data for a downstream task. The local autoencoders are trained for each modality and aggregated at the central server to obtain a global autoencoder using a multi-modal FedAvg algorithm. In \cite{xiong2022unified}, a unified framework is proposed to integrate the complementary information from different modalities using co-attention mechanism. The modality-specific backbones are used for the local data associated with different modalities on clients. During the aggregation phase, the parameters of modality-specific backbones and co-attention are shared to the central server. In \cite{chen2022fedmsplit}, a multi-modal FL algorithm (denoted as FedMSplit) is proposed to capture the correlations between different data modalities by using a dynamic graph structure for the scenario where some modalities might be missing. In \cite{chen2024feddat}, the federated dual-adapter teacher algorithm (denoted as FedDAT) is introduced to improve the adaptation and performance of foundation models in FL settings that involve diverse modalities and data distributions. To this end, this algorithm uses a pre-trained foundation model with parameter-efficient fine-tuning adapters to minimize computational overhead. Additionally, the dual-adapter teacher model faciliates in handling data heterogeneity by regulating client updates and promoting effective knowledge transfer. To the best of our knowledge, there is no work in the RS literature on multi-modal FL.

\begin{figure*}[t]
\centering
\centerline{\includegraphics[width=2\columnwidth,keepaspectratio]{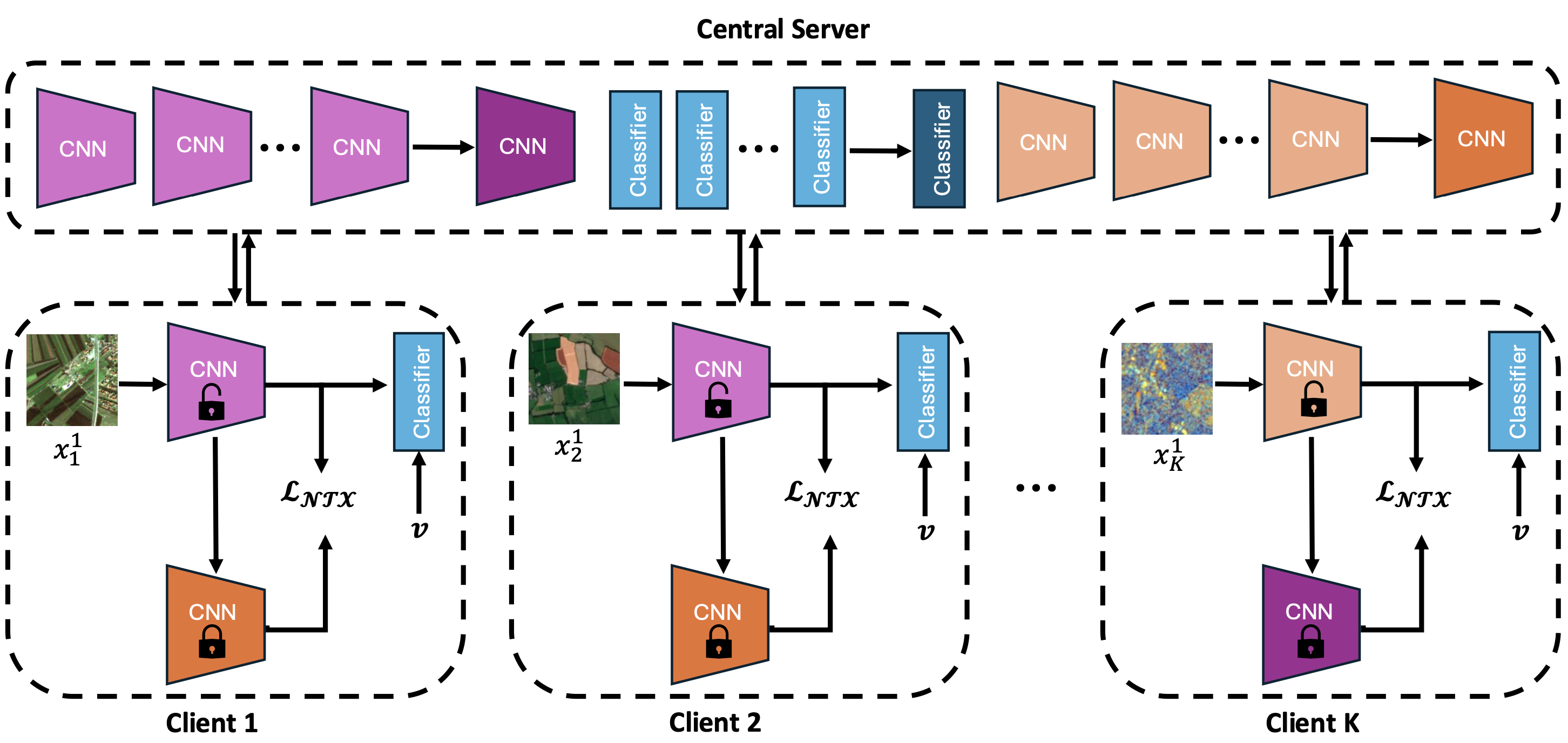}}
\caption{An illustration of our framework. For the sake of simplicity, it is assumed that the images distributed across clients are associated with two different data modalities.  }
\label{figure:module2}
\end{figure*}

\section{Proposed Multi-modal Federated Learning Framework}

Let $K$ denotes the total number of clients and $C_i$ represents the $i$th client for $1 \leq i \leq K$. Let $D_i = \{(\boldsymbol{x}_{i}^z, \boldsymbol{y}_{i}^z)\}_{z=1}^{M_i}$ be the corresponding local training set of $C_i$, where  $M_i$ is the number of images in $D_i$ and $z \in \{1, \ldots, M_i\}$. $\boldsymbol{x}_{i}^z \in \mathbb{R}^{H_i \times W_i \times V_i}$ represents the $z$th RS image in $D_i$, where $H_i$ and $W_i$ refer the height and width, respectively, while $V_i$ represents the number of channels, which can vary across clients depending on the modality of the data. The corresponding image annotation is denoted by $\boldsymbol{y}_{i}^z$. We assume that data modalities can vary among clients (i.e., the data modalities associated with \(x_{i}^{z}\) and \(x_{j}^{z}\) can be different for \(i \ne j\)). Annotations for the training set can be provided at scene level or pixel level in the context of RS image classification. In the case of scene-level annotations, each training image is labeled with either multiple labels (multi-labels) or a single label related to the content of the entire scene. In the case of pixel-level annotations, a class label is assigned to each pixel. The proposed framework aims to learn DL model parameters from unshared, multi-modal training images distributed across clients without requiring access to the training images on clients while addressing the limitations of training data heterogeneity. To this end, our framework comprises three key modules: 1) multi-modal fusion (MF); 2) feature whitening (FW); and 3) mutual information maximization (MIM). The MF module extracts modality-specific features while ensuring privacy-preserving decentralized learning. These extracted features are then aligned using the FW module, which reduces feature distribution discrepancies across different clients by whitening the feature representations. Finally, the MIM module enhances inter-client consistency by maximizing mutual information across modalities, ensuring that the learned feature representations remain modality-agnostic for global model aggregation. Fig. \ref{figure:module2} shows an illustration of our framework. In the following, we first provide general information on FL, and then present the modules of our framework in detail.

\subsection{Basics on FL}

Let $\phi_i$ and $\mu_i$ be the image encoder and task head of the local DL model in $C_i$, respectively, and let $w_i$ denote the set of parameters of the local model. The task head may consist of fully connected or convolutional layers, depending on the learning task. The local model provides the class probabilities $\boldsymbol{r}_{i}^{z}$ associated with $\boldsymbol{x}_{i}^{z}$. The local model is trained on $D_i$ for $E$ epochs using the cross-entropy (CE) loss function $\mathcal{L}_{\text{CE}}$, which can be either categorical CE (for pixel-based and scene-level single-label image classification) or binary CE (for scene-level multi-label image classification). The goal for $C_i$ is to find the optimal local model parameters $w_i^*$ by minimizing the corresponding local objective $\mathcal{O}_i$, which is an empirical risk minimization. Then, the optimal global model parameters $w^*$ can be obtained by minimizing the global objective function, which is a weighted sum of all the local objectives. These can be formulated as follows:

\begin{equation}
\begin{aligned}
\mathcal{O}_i (\mathcal{B};w_i) &=\!\!\!\!\!\!\! \sum_{(\boldsymbol{x}_i^z,\boldsymbol{y}_i^z)\in \mathcal{B}}\!\!\!\!\mathcal{L}_{\text{CE}}(\mu_i(\phi_i(\boldsymbol{x}_{i}^z;w_i)), \boldsymbol{y}_{i}^{z}),\\ 
w_i^* &= \arg \min_{w_i} \mathcal{O}_i(D_i;w_i).
\end{aligned}
\end{equation}
The goal of FL is to find the optimal global model parameters $w^*$ by minimizing the global objective function, which is a weighted sum of all the local objectives as follows: 
\begin{equation}
w^* = \arg \min_{w} \sum_{i=1}^{K} \alpha_i \mathcal{O}_{i}(D_i;w),
\label{minimization}
\end{equation}
where  $\alpha_i$ represents a hyperparameter that controls the significance of client's parameters for aggregation. To this end, the parameters of the global model $w$ are updated through the aggregation of local model parameters $w_i$ of $C_i$ at each FL round as follows:
\begin{equation}
\begin{aligned}
w &= \sum_{i=1}^{K} \alpha_i w_i,
\label{fedavg}
\end{aligned}
\end{equation}
In most FL algorithms, $\alpha_i$ is proportional to the quantity of data available on each client (i.e., $\alpha_i= \frac{\lvert D_i \rvert}{\lvert D \rvert}$, where $D=\bigcup_ {i \in \{ 1, 2,..., K \} } D_i$ represents the whole training set). The process of local training and parameter aggregation iterates sequentially multiple times until the convergence of the global model, which is determined by the number $R$ of communication rounds.

\subsection{Multi-Modal Fusion (MF) Module}

To collaboratively leverage data from different modalities during the inference phase in multi-modal learning, the feature vectors of the images associated with different modalities of the same geographical area need to be extracted using the modality-specific backbones. It ensures that the unique characteristics and complementary information provided by each sensor type are accurately modeled and integrated into a cohesive representation \cite{sun2023single}. However, in multi-modal FL, the images from different modalities might be stored under different clients. Therefore, it is not possible to directly access the images from other clients due to data privacy and decentralization constraints. The MF module aims to perform FL through iterative model averaging on decentralized and unshared multi-modal RS training sets by defining modality-specific backbones and a common classifier. Therefore, our framework can operate without direct access to the raw images from other clients for feature fusion and aggregation. At the central server, the parameters of image encoders are aggregated according to the modalities (i.e., the clients with the same backbones are aggregated among themselves). Let $P$ be the total number of modalities, $\psi_m$ be a modality-specific backbone $\{\psi_m\}_{m=1}^P$ while $1\leq m\leq P \leq K$ and $\theta$ be a common task head. To collaboratively leverage data from different modalities during the inference phase, the feature vectors are fused with zero vector $\boldsymbol{v}$ = $\vec{0}$ through concatenation to mimic the inference phase and fed into $\mu_i$. Therefore, the local objective function is updated as follows:

\begin{equation}
    \mathcal{O}_i (\mathcal{B};w_i) \!=\!\!\!\!\!\!  \sum_{\boldsymbol{x}_i^z, \boldsymbol{y}_i^z \in \mathcal{B}} \!\!\!\!\! \mathcal{L}_{CE}(\theta ( \phi_i(\boldsymbol{x}_i^z) \!\! \mathbin\Vert \! \boldsymbol{v}),\boldsymbol{y}_i^z).
\label{moon2}
\end{equation}
Due to this module, we can define different architectures for different modalities and the iterative model averaging can be employed when the clients are associated with different data modalities. By enabling modality-specific learning and aggregation, the MF module ensures that the unique characteristics of each modality are captured, leading to a more comprehensive and generalized model that can effectively handle multi-modal RS data. By this way, data privacy is maintained while still enabling the model to benefit from the rich, multi-modal information distributed across various clients.

\subsection{Feature Whitening (FW) Module}

When training data is non-IID among clients, the aggregation of model parameters often leads to a global model that diverges from its optimal performance \cite{huba2022papaya}. This divergence is particularly evident when the data in clients is associated with different modalities, further increasing heterogeneity and thus reducing the global model's ability to generalize across diverse data \cite{chen2021communication}. To address this limitation, the FW module aims to overcome the limitations of training data heterogeneity across clients, when clients include images from different data modalities. The FW module employs batch whitening layers as opposed to traditional batch normalization layers, inspired by the method introduced in \cite{roy2019unsupervised}. The batch whitening layer $BW$ for $\boldsymbol{x}_{z,p}^i$ is defined as follows:
\begin{equation}
\begin{split}
\boldsymbol{\hat{x}}^z_{i} &= W_B (\boldsymbol{x}^z_{i} - \mu_B),\\
BW ({\boldsymbol{x}}_{i,p}^z) &= \gamma_p \boldsymbol{\hat{x}}^z_{i,p} + \beta_p,
\end{split}
\label{fw}
\end{equation}
where ${\boldsymbol{x}}_{z,p}^i$ is the $p$th element of ${\boldsymbol{x}}_{z}^i$, $\mu_B$ is the mean vector, $\gamma_p$ is the scaling parameter and $\beta_p$ is the shifting parameter. This module uses domain-specific alignment layers to compute the covariance matrices of intermediate features. Then, the data distribution differences between clients can be aligned using the covariance matrices. By transforming the feature vectors into a whitened space where they are decorrelated and have unit variance, the FW module effectively reduces the impact of data heterogeneity, leading to a more stable and generalizable global model. In multi-modal FL, the batch whitening layers offer a critical advantage over batch normalization layers by more effectively addressing training data heterogeneity. While batch normalization layers standardize features to have zero mean and unit variance, they do not address correlations between features, which is an important factor when integrating information from multiple modalities. The batch whitening layers, on the other hand, transform features to have zero mean and unit variance while also decorrelating them, resulting in a covariance matrix that is an identity matrix. This decorrelation ensures that features from different clients and modalities are similarly distributed. In detail, the effectiveness of the FW module in FL can be analyzed through its role in statistical decorrelation and information disentanglement. The FW module enforces independence by removing linear correlations between features, which helps in stabilizing feature representations across diverse client distributions. In non-IID settings, traditional feature normalization methods (e.g., batch normalization) fail to account for feature dependencies, leading to shifts in learned representations. The FW module, however, guarantees that the transformed features maintain a spherical distribution with unit variance, reducing the risk of model bias towards dominant client distributions. As a result, the model becomes more robust to varying data distributions, improving its overall performance. By transforming features into a common spherical distribution, the batch whitening layers mitigate the impact of local data peculiarities, leading to a global model that generalizes better across different characteristic of data on clients \cite{xu2023personalized}. This enhanced generalization is crucial in multi-modal FL, where the global model must perform well across diverse and unseen data distributions. Moreover, the batch whitening layers contribute to more stable training by ensuring that intermediate feature representations from various clients are more aligned. This alignment prevents the divergence issues often encountered with non-IID data when using batch normalization layers.

\subsection{Mutual Information Maximization (MIM) Module}

In multi-modal learning, accurate alignment of representations across different modalities is crucial. Each modality typically has its own unique feature space, which can lead to difficulties in learning a cohesive and unified representation of the data \cite{quan2023novel}. A widely used approach to address this problem is to employ contrastive learning loss, which maximizes the similarity between different modalities \cite{sumbul2022novel}. In that approach, the images captured from the same geographical area but obtained through different modalities are defined as positive pairs. However, it is not possible to apply this approach when there is no access to images in different modalities from the same geographical area. The MIM module aims to model the mutual information between different modalities by maximizing the similarity without accessing the images from different modalities. In order to accomplish this, we use the NT-Xent loss function $\mathcal{L}_{NTX}$ \cite{chen2020simple}, which corrects the local updates by maximizing the mutual information of representation learned by the current local model and the representation learned by the global model obtained from the clients with different data modalities. In the proposed framework, the features of an image extracted from the local model and those extracted from the global model of other modalities are denoted as positive pairs (e.g., optical image features and radar image features of the same region). To find the negative pairs, we apply the same process with different images in the batch. The feature vectors are obtained by feeding images into both local and global models. Since the global models are not trained on images of the same modality as those in the local client, direct input to the global model is not feasible. Thus, images are fed into the initial convolutional layer of the local model, and the resulting output is then forwarded to the global models. Then, the $\mathcal{L}_{NTX}$ for a given mini-batch $\mathcal{B}$ is calculated as follows:

\begin{equation}
\mathcal{L}_{NTX}(\mathcal{B}) \! = \!\! - \!\!\! \sum_{\boldsymbol{x}_i^z \in \mathcal{B}} \!\!  \log \frac{ e^{S(  \phi_i(\boldsymbol{x}_i^z) ,  \psi_m(\boldsymbol{x}_i^z)  ) / \tau } } {\sum\limits_{\boldsymbol{x}_i^t \in \mathcal{B}} \mathbbm{1}_{[z \neq t]} e^{S(  \phi_i(\boldsymbol{x}_i^z) ,  \psi_m(\boldsymbol{x}_i^t)  ) / \tau } },
\end{equation}
where $\tau$ is a temperature parameter, $\mathbbm{1}$ is the indicator function, $S(.,.)$ measures cosine similarity and $\mathbin\Vert$ is the concatenation operator. Accordingly, we define the local objective function based on $\mathcal{L}_{NTX}$ and $\mathcal{L}_{CE}$ as follows:
\begin{equation}
    \mathcal{O}_i (\mathcal{B};w_i) \!=\! \mathcal{L}_{NTX}\! (\mathcal{B}) \!+\!\!\!\!\!\!  \sum_{\boldsymbol{x}_i^z, \boldsymbol{y}_i^z \in \mathcal{B}} \!\!\!\!\! \mathcal{L}_{CE}(\theta ( \phi_i(\boldsymbol{x}_i^z) \!\! \mathbin\Vert \! \boldsymbol{v}),\boldsymbol{y}_i^z).
\label{moon1}
\end{equation}
In multi-modal FL, local models may develop highly domain-specific representations that hinder global aggregation. By introducing MIM module, the proposed framework regularizes feature extraction across different modalities, ensuring that local updates retain essential modality-agnostic properties. Through MIM module, local models effectively learn shared representations across different modalities, improving the robustness and adaptability of the global model during aggregation. Furthermore, its integration into FL enhances inter-client alignment, leading to a more generalizable model in RS applications.


\begin{table*}[t]
\renewcommand{\arraystretch}{1.5}
\setlength\tabcolsep{10pt} 
\caption{$F_1$ Scores (\%), local training complexity (in seconds), and FLOPs obtained by our framework and the other FL algorithms under decentralization scenarios for the BigEarthNet-MM archive.}
\label{tab:DS_1}
\centering
\begin{tabular}{@{}lccccccc@{}}
\hline
\textbf{Algorithms} & \multicolumn{5}{c}{$\boldsymbol{F_1}$ \textbf{Scores}} & \textbf{ \thead[c]{Local Training\\Complexity} } & \textbf{ \thead[c]{FLOPs} }\\ \cline{2-6}
 & \textbf{DS1-BEN} & \textbf{DS2-BEN} & \textbf{DS3-BEN} & \textbf{DS4-BEN} & \textbf{DS5-BEN} &  \\ \hline
FedAvg~\cite{FedAvg} & 73.68 & 49.71 & 47.97 & 49.12 & 47.79 & \textbf{128} & \textbf{1$\times$} \\ 
SCAFFOLD~\cite{karimireddy2020scaffold} & 76.92 & 50.07 & 48.72 & 50.80 & 50.43 & 132 & 1.2$\times$ \\ 
MOON~\cite{li2021model} & 76.70 & 52.74 & 48.83 & 49.97 & 51.51 & 168 & 1.4$\times$\\ 
FedDC~\cite{gao2022feddc} & 76.81 & 51.43 & 48.37 & 50.05 & 52.14 & 136& 1.2$\times$ \\ 
Our Framework & \textbf{78.48} & \textbf{61.12} & \textbf{60.78} & \textbf{58.27} & \textbf{60.30} & 194& 1.5$\times$ \\ \hline
\end{tabular}
\end{table*}

\section{Data Set Description And Design of Experiments}

\subsection{Data Set Description}

The experiments were conducted on the BigEarthNet-MM v1.0 \cite{sumbul2021bigearthnet} and the Dynamic World-Expert \cite{fuller2024croma} benchmark archives described in the following.

\subsubsection{BigEarthNet-MM Archive}

The BigEarthNet-MM v1.0 comprises 590,326 multi-modal image pairs. Each image pair includes Sentinel-1 and Sentinel-2 images acquired on the same geographical area. These pairs are collected from 10 European countries. In this paper, we utilized a subset of BigEarthNet-MM, including images from 7 countries (Austria, Belgium, Finland, Ireland, Lithuania, Serbia, Switzerland). Each Sentinel-2 image consists of 120$\times$120 pixels for 10m bands, 60$\times$60 pixels for 20m bands, and 20$\times$20 pixels for 60m bands. The 60m bands were excluded from our experiments, and the 20m bands were resized using bicubic interpolation to 120×120 pixels, resulting in 10 bands per image. The images were annotated with multi-labels based on the CORINE Land Cover Map database. We used the 19-class nomenclature introduced in \cite{sumbul2021bigearthnet}. We used the official training and testing splits of BigEarthNet-MM proposed in \cite{sumbul2021bigearthnet}, and defined five decentralization scenarios defined as follows:

\begin{itemize}
\item Decentralization Scenario 1 - BigEarthNet-MM (DS1-BEN): The images acquired during summer were randomly distributed among different clients.
\item Decentralization Scenario 2 - BigEarthNet-MM (DS2-BEN): The images acquired during summer were distributed based on countries and each client has images only from one country.
\item Decentralization Scenario 3 - BigEarthNet-MM (DS3-BEN): The images were distributed based on countries independently from the acquisition date and each client has images from only one country.
\item Decentralization Scenario 4 - BigEarthNet-MM (DS4-BEN): The images acquired during the summer were randomly distributed among different clients. Additionally, we used only half of the available Sentinel-2 images to evaluate the effectiveness of our framework in the presence of missing Sentinel-2 images.
\item Decentralization Scenario 5 - BigEarthNet-MM (DS5-BEN): The images acquired during the summer were randomly distributed among different clients. Additionally, we used only half of the available Sentinel-1 images to evaluate the effectiveness of our framework in the presence of missing Sentinel-1 images.
\end{itemize}

For each scenario, we include 14 clients in total, from which 7 clients contain only Sentinel-1 images, whereas the remaining ones consist of only Sentinel-2 images. All the defined clients participate in each round of the local training. We would like to note that the considered scenarios are associated to varying levels of training data heterogeneity across clients. In the first three scenarios, all image pairs from the same geographical area are available to the clients. Among these, DS1-BEN is associated to the lowest heterogeneity, as images from the same season are randomly assigned to clients. DS3-BEN is associated to the highest data heterogeneity, as it includes images from various seasons and countries. 


\subsubsection{Dynamic World-Expert Archive}

The Dynamic World-Expert contains 20,422 multi-modal image pairs, each of which includes Sentinel-1 and Sentinel-2 images acquired on the same geographical area. Each image consists of 96$\times$96 pixels. We used the official training and testing splits of Dynamic World-Expert presented in \cite{fuller2024croma} and defined three decentralization scenarios described as follows: 

\begin{itemize}
\item Decentralization Scenario 1 - Dynamic World-Expert (DS1-DWE): The images acquired during the summer were randomly distributed among different clients.
\item Decentralization Scenario 2 - Dynamic World-Expert (DS2-DWE): The images acquired during the summer were randomly distributed among different clients. Additionally, we used only half of the available Sentinel-2 images to evaluate the effectiveness of our framework in the presence of missing Sentinel-2 images.
\item Decentralization Scenario 3 - Dynamic World-Expert (DS3-DWE): The images acquired during the summer were randomly distributed among different clients. Additionally, we used only half of the available Sentinel-1 images to evaluate the effectiveness of our framework in the presence of missing Sentinel-1 images.
\end{itemize}

For each scenario, we include 14 clients in total, from which 7 clients contain only Sentinel-1 images, whereas the remaining ones consist of only Sentinel-2 images. All the defined clients participate in each round of the local training. Among these, DS1-DWE is associated to the lowest heterogeneity, as the images associated with the same geographical location are present in both modalities. DS2-DWE and DS3-DWE have similar levels of heterogeneity due to an equal number of images with missing modalities.

\subsection{Design of Experiments}

We carried out different kinds of experiments in order to: 1) conduct a sensitivity analysis; 2) compare our framework with the state-of-the-art FL algorithms; and 3) achieve an ablation study. We tested our framework in the context of two learning tasks: 1) multi-label scene classification using BigEarthNet-MM; and 2) pixel-based classification using Dynamic World-Expert. The results are provided as $F_1$ score (for multi-label classification), overall accuracy (for pixel-based classification), and training complexity (for both learning tasks). We selected the following state-of-the-art algorithms: i) FedAvg~\cite{FedAvg}; ii) SCAFFOLD~\cite{karimireddy2020scaffold}; iii) MOON~\cite{li2021model}; and iv) FedDC~\cite{gao2022feddc}. However, it is not feasible to directly use them. Thus, we adapted them by using the algorithms for each modality during training, followed by a late fusion technique during inference. We utilized the ResNet-50 architecture as a backbone of the global model. We set the number of communication rounds to 40. Each client employed the same DL architecture as the global model for their local model except for the first convolutional layer (the first layer needs to be updated due to the varying input size of the training data). During each communication round, local models were trained for one epoch with the mini-batch size set of 512. For local model training, the Adam variant of stochastic gradient descent was used with an initial learning rate of $10^{-3}$ and a weight decay of 0.9.

\section{Experimental Results}

\subsection{Sensitivity Analysis of the Proposed Framework}

We performed the sensitivity analysis of the proposed framework under different number of clients. Fig. 2 shows the $F_1$ scores over multiple communication rounds under DS1-BEN when different numbers $K$ of clients are used in the proposed framework. One can observe from the figure that as $K$ increases, our framework leads to a decrease in $F_1$ scores. As an example, the $F_1$ score of the proposed framework decreases by 1.5\% when $K=7 \rightarrow K=14$, while it decreases by 5.2\% when $K=14 \rightarrow K=28$. This is primarily because each client receives fewer training samples, resulting in limited local learning and poorer generalization. Additionally, the increased number of clients leads to higher statistical heterogeneity, causing variations in local model updates that can negatively impact global model convergence. As shown in the results, while $K = 7$ achieves the highest $F_1$ score, it involves fewer clients, which may not be ideal for simulating real-world FL scenarios with a larger number of participants. On the other hand, $K = 28$ leads to the lowest $F_1$ score, which negatively impacts the global model convergence. We obtained similar conclusions from the experiments conducted on Dynamic World-Expert dataset, though the results are not included due to space limitations. For the rest of the experiments, we set the number K of clients as $K=14$. By selecting $K = 14$, the experiments maintain a reasonable level of data distribution per client, ensuring that local models have enough training samples to learn effectively while still including a sufficient number of clients to represent decentralized learning conditions.

\begin{figure}[t]
    \centering
    \begin{tikzpicture}[scale = 0.8]%
	\begin{axis}[
    legend columns=1,
    height=7cm,
    width=10.8cm,
    legend style={font=\scriptsize, at={(axis cs:32,62)},anchor=south west},
    grid=both,
    grid style={line width=.1pt, draw=gray!10},
    major grid style={line width=.2pt,draw=gray!50},
    minor x tick num=3,
    minor y tick num=4,
    xlabel= {\small Communication Round},
    ylabel= {\small $F_1$ Score (\%)},
    xmin=3,xmax=40,
    ymin=60, ymax=81]
    \addplot+[name path=capacity,dashed,color=orange,mark=none,line width=1pt] table [x=comm_rounds, y=client-7, col sep=comma] {sensitivity_analysis.csv};\addlegendentry{$K=7$};
    \addplot+[name path=capacity,solid,color=cyan,mark=none,line width=1pt] table [x=comm_rounds, y=client-14, col sep=comma] {sensitivity_analysis.csv};\addlegendentry{$K=14$};
    \addplot+[name path=capacity,densely dotted,color=magenta,mark=none,line width=1pt] table [x=comm_rounds, y=client-28, col sep=comma] {sensitivity_analysis.csv};\addlegendentry{$K=28$}
	\end{axis}
    \end{tikzpicture}
\caption{$F_1$ score versus communication round obtained by the proposed framework when the different numbers $K$ of clients are considered under DS1-BEN.}
\label{fig:sens}
\end{figure}
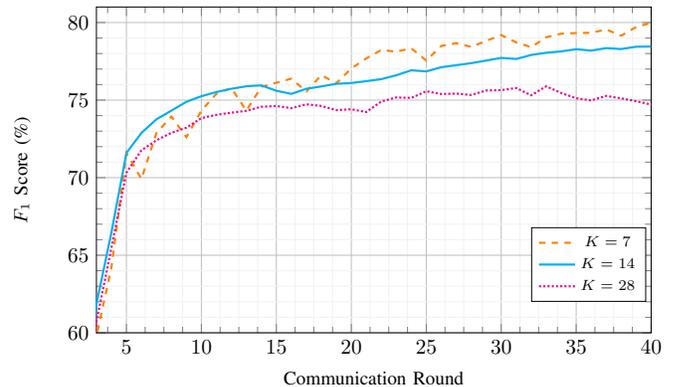

\subsection{Comparison of Our Framework with State-of-the-art FL Algorithms}

In this subsection, we compare our framework with the state-of-the-art FL algorithms in the context of both multi-label scene classification (MLC) and pixel-based classification problems.
\vspace{0.6em}
\subsubsection{Multi-Label Scene Classification}

In the first set of trials, we compare our framework with FL algorithms when multi-modal images associated with the same geographical area are available during the inference phase in the context of multi-label scene classification. Table \ref{tab:DS_1} shows the corresponding results on BigEarthNet-MM archive. As one can see from the table that as data heterogeneity among clients increases through our decentralization scenarios, the performance of each algorithm declines in terms of $F_1$ scores. For instance, our framework achieves over 17.36\% and 17.70\% higher scores in DS1-BEN compared to DS2-BEN and DS3-BEN, respectively. This indicates that higher levels of data heterogeneity pose significant challenges for model generalization across different client distributions. One can also observe from the table that our framework consistently provides the highest $F_1$ scores under all scenarios. As an example, our framework outperforms FedAvg, SCAFFOLD, MOON, and FedDC by 4.80\%, 1.56\%, 1.78\%, and 1.67\%, respectively, under DS1-BEN. One can also assess from the table that the differences in the performance are higher with the higher levels of non-IID scenarios (e.g., DS2-BEN and DS3-BEN) compared to the lowest non-IID scenario (i.e., DS1-BEN). As an example, compared to the MOON algorithm under DS2-BEN and DS3-BEN, our framework provides 8.38\% and 11.90\% higher $F_1$ scores, which are significantly higher than those under DS1-BEN. This indicates that our framework more effectively leverages and integrates diverse data modalities than the other algorithms, resulting in superior MLC accuracy. To observe the effectiveness of our framework and other FL algorithms under scenarios with missing modalities, we obtained the results also under DS4-BEN and DS5-BEN. According to the results, we obtained $F_1$ scores of 58.27\% and 60.30\% using our framework under DS4-BEN and DS5-BEN, which are 8.22\% and 8.16\% higher, respectively, than those obtained by the FedDC algorithm. These results prove that the our framework performs better than the other state-of-the-art algorithms, when modalities associated with the some geographical areas are missing. This is because our framework is adjusted to the absence of specific modalities due to the optimized learning process among clients. This ensures high performance even when some data modalities are missing. Table \ref{tab:DS_1} shows the local training complexity and FLOPs for the BigEarthNet-MM archive. By analysing Table \ref{tab:DS_1}, one can also observe that the average completion time of a training round on one client increases by more than 20\% when using the proposed framework compared to FedAvg, SCAFFOLD, MOON, and FedDC. This shows that our framework outperforms other FL algorithms in terms of accuracy with a cost of slight increase in local training complexity. Moreover, our framework offers the best trade-off by significantly improving $F_1$ scores at a reasonable increase in FLOPs. Fig. 3 presents an example of multi-modal BigEarthNet-MM image pairs with the true multi-labels and the multi-labels assigned by the FedAvg, SCAFFOLD, MOON, FedDC, and the proposed framework. As one can see from the figure, the proposed framework accurately predicts all relevant classes without introducing any incorrect labels. In contrast, SCAFFOLD and FedDC misclassify images by assigning unrelated labels. As an example, in the third image, both SCAFFOLD and FedDC incorrectly predict the presence of urban fabric, despite these classes being absent from the image.

In the second set of trials, we compare our framework with FL algorithms when only one modality from a specific geographical area is available during the inference phase. Table \ref{tab:DS_3} shows the results obtained using only Sentinel-1 images (denoted as BEN-S1) or Sentinel-2 images (denoted as BEN-S2) during the inference phase on the BigEarthNet-MM archive under DS4-BEN and DS5-BEN. 
One can observe from the table that our framework outperforms other FL algorithms under DS4-BEN and DS5-BEN. As an example, our framework provides 19.10\% and 19.65\% higher $F_1$ scores under DS4-BEN and DS5-BEN, respectively, compared to the MOON algorithm when using only BEN-S1 during the inference. As an another example, our framework provides 3.29\% and 3.04\% higher $F_1$ scores under DS4-BEN and DS5-BEN, respectively, compared to the FedDC algorithm when using only BEN-S2 during the inference. The significantly higher $F_1$ score achieved with BEN-S1 using our framework is due to the fact that local models trained on Sentinel-1 benefit from the Sentinel-2 global model during the learning process.

\begin{table}[t]
\renewcommand{\arraystretch}{1.6}
\setlength\tabcolsep{5pt}
\caption{$F_1$ Scores (\%) obtained by our framework and the other FL algorithms under DS4-BEN and DS5-BEN for the BigEarthnet-MM archive. BEN-S1 denotes Sentinel-1 images in the BigEarthnet-MM. BEN-S2 denotes Sentinel-2 images in the BigEarthnet-MM}
\label{tab:DS_3}
\centering
\begin{tabular}{@{}lccccc@{}}
\hline
\textbf{Algorithm} & \multicolumn{2}{c}{\textbf{DS4-BEN}} & \multicolumn{2}{c}{\textbf{DS5-BEN}} \\ 
& \textbf{BEN-S1} & \textbf{BEN-S2} & \textbf{BEN-S1} & \textbf{BEN-S2} \\ \hline
FedAvg~\cite{FedAvg} & 28.53 & 54.89 & 27.07 & 55.78 \\ 
SCAFFOLD~\cite{karimireddy2020scaffold} & 37.48 & 57.09 & 33.70 & 56.04 \\ 
MOON~\cite{li2021model} & 35.59 & 57.21 & 33.28 & 58.98 \\ 
FedDC~\cite{gao2022feddc} & 38.61 & 56.57 & 37.87 & 58.45 \\ 
Our Framework & \textbf{54.69} & \textbf{59.86} & \textbf{52.93} & \textbf{61.49} \\ \hline
\end{tabular}%
\end{table}

\begin{figure*}[t]
\centering
\centerline{\includegraphics[width=2\columnwidth,keepaspectratio]{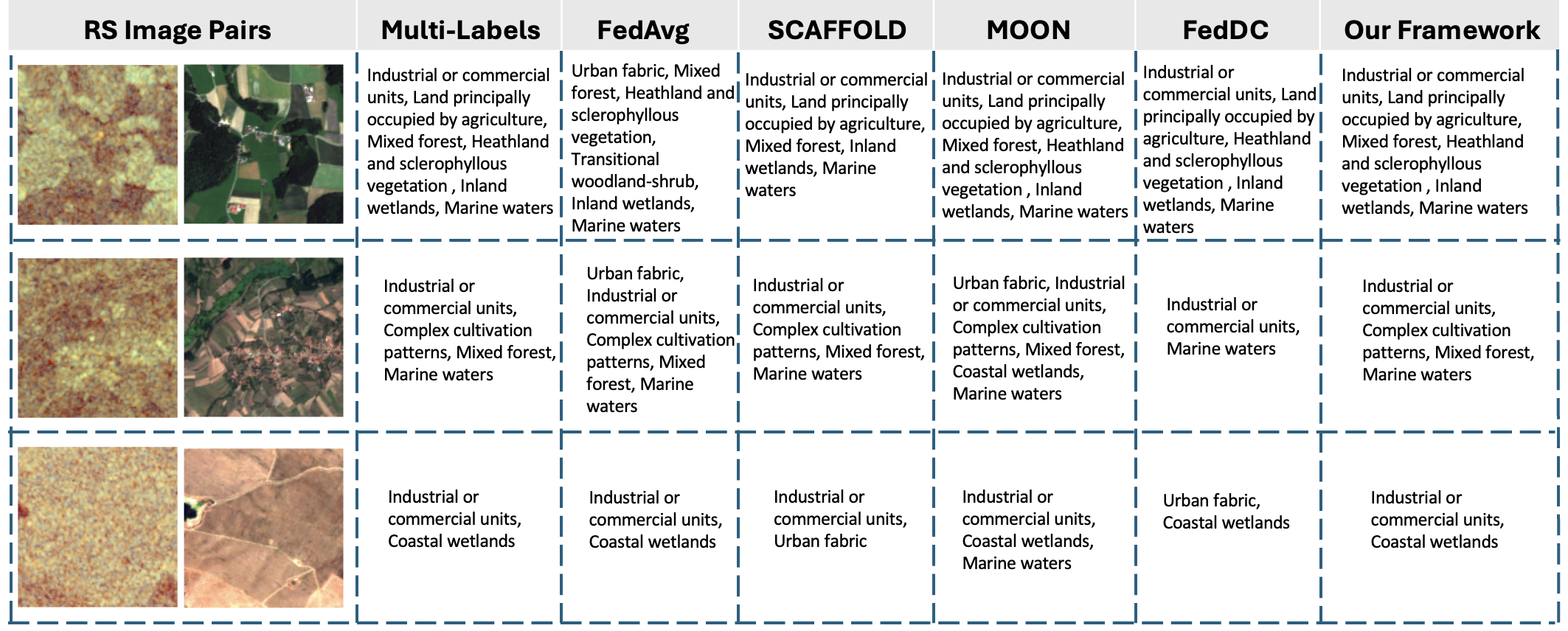}}
\caption{An example of BigEarthNet-MM image pairs with the true multi-labels and the multi-labels assigned by the FedAvg, SCAFFOLD, MOON, FedDC, and the proposed framework.  }
\label{example}
\end{figure*}

\subsubsection{Pixel-Based Classification}

In the first set of trials, we compare our framework with FL algorithms when multi-modal images associated with the same geographical area are available during the inference phase in the context of pixel-based classification. Table \ref{tab:DS_2} shows the corresponding results on Dynamic World-Expert archive. As one can also see from the table that, our framework outperforms other FL algorithms across every scenario. As an example, we obtained the $F_1$ scores of 51.42\%, 48.14\%, and 50.76\% using our framework under DS1-DWE, DS2-DWE, and DS3-DWE, which are 1.16\%, 0.28\% and 0.48\% higher, respectively, than those obtained by the FedDC algorithm. This indicates that our framework demonstrates superior performance in pixel-based classification tasks as in the multi-label classification tasks. However, the performance differences between our framework and other algorithms on the Dynamic World-Expert archive are not as high as those observed with BigEarthNet-MM. This is because the non-IID level in the decentralization scenarios of the Dynamic World-Expert archive is lower than that of BigEarthNet-MM. One can observe from the table that the training complexity (i.e., average completion time of a training round) of our framework is higher than that of other FL algorithms. As an example, the local training complexity of our framework is 28\% higher than that of the FedDC algorithm under all scenarios. This complexity primarily arises due to the MIM module, as it requires computing the similarity between features from different data modalities.

In the second set of trials, we compare our framework with FL algorithms when only one modality from a specific geographical area is available during the inference phase. Table \ref{tab:DS_4} shows the results obtained using only Sentinel-1 images (denoted as DWE-S1) or Sentinel-2 images (denoted as DWE-S2) during the inference phase on the Dynamic World-Expert archive under DS2-DWE and DS3-DWE. As one can observe from the table that our framework outperforms other FL algorithms under DS2-DWE and DS3-DWE. As an example, our framework provides 1.96\% and 1.62\% higher $F_1$ scores under DS2-DWE and DS3-DWE, respectively, compared to the MOON algorithm when using only DWE-S1 during the inference. As an another example, our framework provides 2.19\% and 0.67\% higher $F_1$ scores under DS2-DWE and DS3-DWE, respectively, compared to the FedDC algorithm when using only DWE-S2 during the inference. These results further prove that our framework exhibits superior performance compared to other algorithms in the case of missing modalities during the inference.

\begin{table}[t]
\renewcommand{\arraystretch}{1.4}
\setlength\tabcolsep{4pt}
\caption{$F_1$ Scores (\%) and local training complexity (in seconds) obtained by our framework and the other FL Algorithms under decentralization scenarios for the Dynamic World-Expert archive.}
\label{tab:DS_2}
\centering
\begin{tabular}{@{}lcccc@{}}
\hline
\textbf{Algorithm} & \multicolumn{3}{c}{$\boldsymbol{F_1}$ \textbf{Scores}} & \textbf{ \thead[c]{Local\\ Training\\Complexity}  } \\ \cline{2-4}
 & \textbf{DS1-DWE} & \textbf{DS2-DWE} & \textbf{DS3-DWE} &  \\ \hline
FedAvg~\cite{FedAvg} & 46.36 & 45.20 & 47.16 & \textbf{58} \\ 
SCAFFOLD~\cite{karimireddy2020scaffold} & 47.44 & 46.07 & 47.89 & 60 \\
MOON~\cite{li2021model} & 49.99 & 47.72 & 49.48 & 76 \\ 
FedDC~\cite{gao2022feddc} & 50.26 & 47.86 & 50.28 & 61 \\ 
Our Framework & \textbf{51.42} & \textbf{48.14} & \textbf{50.76} & 85 \\ \hline
\end{tabular}
\end{table}

\begin{table}[t]
\renewcommand{\arraystretch}{1.6}
\setlength\tabcolsep{5pt}
\caption{$F_1$ Scores (\%) obtained by our framework and the other FL algorithms under DS2-DWE and DS3-DWE for the Dynamic World-Expert archive. DWE-S1 denotes Sentinel-1 images in the Dynamic World-Expert. DWE-S2 denotes Sentinel-2 images in the Dynamic World-Expert}
\label{tab:DS_4}
\centering
\begin{tabular}{@{}lccccc@{}}
\hline
\textbf{Algorithm} & \multicolumn{2}{c}{\textbf{DS2-DWE}} & \multicolumn{2}{c}{\textbf{DS3-DWE}} \\ 
& \textbf{DWE-S1} & \textbf{DWE-S2} & \textbf{DWE-S1} & \textbf{DWE-S2} \\ \hline
FedAvg~\cite{FedAvg} & 43.42 & 43.82 & 42.58 & 44.75 \\ 
SCAFFOLD~\cite{karimireddy2020scaffold} & 44.23 & 47.56 & 45.11 & 48.74 \\ 
MOON~\cite{li2021model} & 45.78 & 45.85 & 45.28 & 45.62 \\ 
FedDC~\cite{gao2022feddc} & 46.85 & 47.12 & 45.56 & 48.19 \\ 
Our Framework & \textbf{47.74} & \textbf{49.31} & \textbf{46.90} & \textbf{48.86} \\ \hline
\end{tabular}%
\end{table}

\subsection{Ablation Study of the Proposed Framework}

In this subsection, we present an ablation study to analyze the effectiveness of each module of our proposed framework. Table \ref{tab:ablation} shows the ${F_1}$ scores obtained by: (i) using only MF module; ii) jointly using MF and FW modules; and iii) jointly using MF and MIM modules under different decentralization scenarios considered for the BigEarthNet-MM in the context of multi-label classification. The results are presented under DS1-BEN, DS2-BEN, and DS3-BEN, while the scenarios with missing modalities (i.e., DS4-BEN and DS5-BEN) are not included. From the table, one can observe that the the ${F_1}$-scores obtained under DS1-BEN are higher than those in DS2-BEN, whereas the scores under DS2-BEN are higher than those under DS3-BEN. As an example, the joint of use of all three modules provides 17.36\% and 17.70\% higher ${F_1}$ scores when compared to the DS2-BEN and DS3-BEN, respectively. This indicates that higher levels of data heterogeneity pose significant challenges for model generalization across different client distributions. From Table \ref{tab:ablation} one can also observe that the joint of use of all three modules results in the highest ${F_1}$ scores compared to other combinations. As an example, under DS2-BEN, the ${F_1}$ score achieved by using all modules is 8.11\% higher than that obtained when only the MF and MIM modules are used. These results prove the effectiveness of reducing data distribution differences and maximizing information consistency between clients. By analysing the table, we can also observe that the joint use of MF and FW modules outperforms those of MF and MIM modules. As an example, the joint use of MF and FW modules provides more than 6\% higher $F_1$ score compared to the joint use of MF and MIM modules under DS2-BEN and DS3-BEN. This indicates that the FW module, by effectively handling non-IID data, contributes to more stable performance and enhanced accuracy in scenarios with high non-IID levels. It is worth noting that the same relative behavior is observed in the results obtained under DS4-BEN and DS5-BEN, as well as when using the Dynamic World-Expert archive in the context of pixel-based classification. In order not to increase the complexity of the paper, we do not report the ablation study for these results.

\begin{table}[t]
\renewcommand{\arraystretch}{1.5}
\caption{$F_1$ scores (\%) obtained by using different combinations of modules in the proposed framework for the BigEarthNet-MM archive.}
\centering
\label{tab:ablation} 
\begin{tabular}{@{}ccc@{\hspace{25pt}}ccc@{}}
\hline
\multicolumn{3}{@{\hspace{-20pt}}c}{\textbf{Modules}} & \multicolumn{3}{c}{\textbf{$\boldsymbol{F_1}$ Scores}} \\ 
\textbf{MF} & \textbf{FW} & \textbf{MIM} & \textbf{DS1-BEN} & \textbf{DS2-BEN} & \textbf{DS3-BEN} \\ 
\hline
\cmark & \xmark & \xmark & 77.68 & 54.85 & 54.12 \\ 
\cmark & \xmark & \cmark & 77.57 & 53.01 & 53.13 \\ 
\cmark & \cmark & \xmark & 78.18 & 60.09 & 59.48 \\ 
\cmark & \cmark & \cmark & \textbf{78.48} & \textbf{61.12} & \textbf{60.78} \\ \hline
\end{tabular}
\end{table}

\section{Conclusion}

In this paper, we present a novel FL framework designed to learn DL model parameters from decentralized and unshared multi-modal RS image archives in the context of RS image classification. Our framework utilizes three modules: 1) multi-modal fusion; 2) feature whitening; and 3) mutual information maximization. The multi-modal fusion module utilizes iterative model averaging to enable learning without directly accessing multi-modal training data on clients. The feature whitening module addresses the challenges of data heterogeneity by aligning data distributions across clients. The mutual information maximization module focuses on capturing mutual information by maximizing the similarity between images from different modalities. Due to the joint use of these modules, our framework strengthens the ability of global model to learn complementary features across diverse data modalities, effectively addressing training data heterogeneity while ensuring that data remains distributed and unshared among clients.

We would like to highlight that our framework is independent from the choice of modality-specific backbones. Consequently, any DL model architecture tailored to a particular modality can be integrated into our framework. This flexibility enables the accurate representation of modality-specific information, enhancing RS image classification performance. In the experiments, we consider two different learning tasks: i) multi-label scene classification using BigEarthNet-MM; and ii) pixel-based classification using Dynamic World-Expert. Experimental result prove the effectiveness of our framework compared to the state-of-the-art FL algorithms in capturing and utilizing multi-modal information for RS image classification under different decentralization scenarios.

We would like to note that our framework is designed under the assumption that each client contains images from a single modality. This assumption aligns with traditional FL for multi-modal data, where each client corresponds to a specific data source. However, extending the framework to scenarios where clients hold multiple modalities would require modifications. One possible approach is to integrate a multi-modal fusion module at the client level, enabling clients to effectively combine different modalities before sharing updates with the server. Techniques such as late fusion, feature alignment strategies, or cross-modal contrastive learning could help mitigate feature conflicts and ensure meaningful knowledge transfer across modalities. As a future work, we will explore these extensions to make the framework more flexible for heterogeneous multi-modal data distributions within each client.


\section*{Acknowledgment}
This work is supported by the European Research Council (ERC) through the ERC-2017-STG BigEarth Project under Grant 759764.

\bibliographystyle{IEEEtran}
{\small
\bibliography{refs}}

\begin{thebibliography}{10}
\providecommand{\url}[1]{#1}
\csname url@samestyle\endcsname
\providecommand{\newblock}{\relax}
\providecommand{\bibinfo}[2]{#2}
\providecommand{\BIBentrySTDinterwordspacing}{\spaceskip=0pt\relax}
\providecommand{\BIBentryALTinterwordstretchfactor}{4}
\providecommand{\BIBentryALTinterwordspacing}{\spaceskip=\fontdimen2\font plus
\BIBentryALTinterwordstretchfactor\fontdimen3\font minus \fontdimen4\font\relax}
\providecommand{\BIBforeignlanguage}[2]{{%
\expandafter\ifx\csname l@#1\endcsname\relax
\typeout{** WARNING: IEEEtran.bst: No hyphenation pattern has been}%
\typeout{** loaded for the language `#1'. Using the pattern for}%
\typeout{** the default language instead.}%
\else
\language=\csname l@#1\endcsname
\fi
#2}}
\providecommand{\BIBdecl}{\relax}
\BIBdecl

\bibitem{lyu2022privacy}
L.~Lyu, H.~Yu, X.~Ma, C.~Chen, L.~Sun, J.~Zhao, Q.~Yang, and S.~Y. Philip, ``Privacy and robustness in federated learning: Attacks and defenses,'' \emph{IEEE Transactions on Neural Networks and Learning Systems}, vol.~35, pp. 8726--8746, 2022.

\bibitem{zhang2022progress}
B.~Zhang, Y.~Wu, B.~Zhao, J.~Chanussot, D.~Hong, J.~Yao, and L.~Gao, ``Progress and challenges in intelligent remote sensing satellite systems,'' \emph{IEEE Journal of Selected Topics in Applied Earth Observations and Remote Sensing}, vol.~15, pp. 1814--1822, 2022.

\bibitem{tuia2023artificial}
D.~Tuia, K.~Schindler, B.~Demir, G.~Camps-Valls, X.~X. Zhu, M.~Kochupillai, S.~D{\v{z}}eroski, J.~N. van Rijn, H.~H. Hoos, F.~Del~Frate \emph{et~al.}, ``Artificial intelligence to advance earth observation: A perspective,'' \emph{IEEE Geoscience and Remote Sensing Magazine}, 2023.

\bibitem{munawar2022remote}
H.~S. Munawar, A.~W. Hammad, and S.~T. Waller, ``Remote sensing methods for flood prediction: A review,'' \emph{MDPI Sensors}, vol.~22, no.~3, pp. 960--969, 2022.

\bibitem{rafiq2023sensordrop}
K.~Rafiq, R.~Appleby, A.~Davies, and B.~Abrahms, ``Sensordrop: A system to remotely detach individual sensors from wildlife tracking collars,'' \emph{Ecology and Evolution}, vol.~13, no.~7, p. e10220, 2023.

\bibitem{yao2022multi}
Y.~Yao, Y.~Zhang, Y.~Wan, X.~Liu, X.~Yan, and J.~Li, ``Multi-modal remote sensing image matching considering co-occurrence filter,'' \emph{IEEE Transactions on Image Processing}, vol.~31, pp. 2584--2597, 2022.

\bibitem{hoffmann2023transformer}
D.~Hoffmann, K.~N. Clasen, and B.~Demir, ``Transformer-based multi-modal learning for multi label remote sensing image classification,'' \emph{IEEE International Geoscience and Remote Sensing Symposium}, 2023.

\bibitem{fang2023classification}
J.~Fang and X.~Yan, ``Classification of multi-modal remote sensing images based on knowledge graph,'' \emph{International Journal of Remote Sensing}, vol.~44, no.~15, pp. 4815--4835, 2023.

\bibitem{weilandt2023early}
F.~Weilandt, R.~Behling, R.~Goncalves, A.~Madadi, L.~Richter, T.~Sanona, D.~Spengler, and J.~Welsch, ``Early crop classification via multi-modal satellite data fusion and temporal attention,'' \emph{MDPI Remote Sensing}, vol.~15, no.~3, pp. 799--823, 2023.

\bibitem{zhang2023meta}
Y.~Zhang, K.~Gong, K.~Zhang, H.~Li, Y.~Qiao, W.~Ouyang, and X.~Yue, ``Meta-{T}ransformer: A unified framework for multimodal learning,'' \emph{arXiv preprint arXiv:2307.10802}, 2023.

\bibitem{moshawrab2023reviewing}
M.~Moshawrab, M.~Adda, A.~Bouzouane, H.~Ibrahim, and A.~Raad, ``Reviewing federated learning aggregation algorithms; strategies, contributions, limitations and future perspectives,'' \emph{Electronics}, vol.~12, no.~10, p. 2287, 2023.

\bibitem{yu2023multimodal}
Q.~Yu, Y.~Liu, Y.~Wang, K.~Xu, and J.~Liu, ``Multimodal federated learning via contrastive representation ensemble,'' \emph{arXiv preprint arXiv:2302.08888}, 2023.

\bibitem{xiong2022unified}
B.~Xiong, X.~Yang, F.~Qi, and C.~Xu, ``A unified framework for multi-modal federated learning,'' \emph{Neurocomputing}, vol. 480, pp. 110--118, 2022.

\bibitem{chen2022fedmsplit}
J.~Chen and A.~Zhang, ``{FedMSplit}: Correlation-adaptive federated multi-task learning across multimodal split networks,'' \emph{ACM SIGKDD Conference on Knowledge Discovery and Data Mining}, pp. 87--96, 2022.

\bibitem{Buyuktas:2023}
B.~Büyüktaş, G.~Sumbul, and B.~Demir, ``Learning across decentralized multi-modal remote sensing archives with federated learning,'' \emph{IEEE International Geoscience and Remote Sensing Symposium}, pp. 4966--4969, 2023.

\bibitem{burgert2022effects}
T.~Burgert, M.~Ravanbakhsh, and B.~Demir, ``On the effects of different types of label noise in multi-label remote sensing image classification,'' \emph{IEEE Transactions on Geoscience and Remote Sensing}, vol.~60, pp. 1--13, 2022.

\bibitem{aksoy2022multi}
A.~K. Aksoy, M.~Ravanbakhsh, and B.~Demir, ``Multi-label noise robust collaborative learning for remote sensing image classification,'' \emph{IEEE Transactions on Neural Networks and Learning Systems}, vol.~35, pp. 6438--6451, 2022.

\bibitem{sumbul2021bigearthnet}
G.~Sumbul, A.~D.~Wall, T.~Kreuziger, F.~Marcelino, H.~Costa, P.~Benevides, M.~Caetano, B.~Demir, and V.~Markl, ``{BigEarthNet-MM}: A large-scale, multimodal, multilabel benchmark archive for remote sensing image classification and retrieval,'' \emph{IEEE Geoscience and Remote Sensing Magazine}, vol.~9, no.~3, pp. 174--180, 2021.

\bibitem{bruzzone2014review}
L.~Bruzzone and B.~Demir, ``A review of modern approaches to classification of remote sensing data,'' \emph{Land Use and Land Cover Mapping in Europe: Practices \& Trends}, pp. 127--143, 2014.

\bibitem{fuller2024croma}
A.~Fuller, K.~Millard, and J.~Green, ``{CROMA}: Remote sensing representations with contrastive radar-optical masked autoencoders,'' \emph{Advances in Neural Information Processing Systems}, vol.~36, 2024.

\bibitem{li2020federated}
T.~Li, A.~K. Sahu, A.~Talwalkar, and V.~Smith, ``Federated learning: Challenges, methods, and future directions,'' \emph{IEEE Signal Processing Magazine}, vol.~37, no.~3, pp. 50--60, 2020.

\bibitem{karimireddy2020scaffold}
S.~P. Karimireddy, S.~Kale, M.~Mohri, S.~Reddi, S.~Stich, and A.~T. Suresh, ``{SCAFFOLD}: Stochastic controlled averaging for federated learning,'' \emph{International Conference on Machine Learning}, pp. 5132--5143, 2020.

\bibitem{gao2022feddc}
L.~Gao, H.~Fu, L.~Li, Y.~Chen, M.~Xu, and C.~Xu, ``{FedDC}: Federated learning with non-iid data via local drift decoupling and correction,'' \emph{IEEE Conference on Computer Vision and Pattern Recognition}, pp. 10\,112--10\,121, 2022.

\bibitem{li2021fedbn}
X.~Li, M.~Jiang, X.~Zhang, M.~Kamp, and Q.~Dou, ``Fed{BN}: Federated learning on non-iid features via local batch normalization,'' \emph{International Conference on Learning Representations}, 2021.

\bibitem{ma2022layer}
X.~Ma, J.~Zhang, S.~Guo, and W.~Xu, ``Layer-wised model aggregation for personalized federated learning,'' in \emph{IEEE/CVF Conference on Computer Vision and Pattern Recognition}, 2022, pp. 10\,092--10\,101.

\bibitem{wu2022communication}
C.~Wu, F.~Wu, L.~Lyu, Y.~Huang, and X.~Xie, ``Communication-efficient federated learning via knowledge distillation,'' \emph{Nature communications}, vol.~13, no.~1, p. 2032, 2022.

\bibitem{zhang2023federated}
X.~Zhang, B.~Zhang, W.~Yu, and X.~Kang, ``Federated deep learning with prototype matching for object extraction from very-high-resolution remote sensing images,'' \emph{IEEE Transactions on Geoscience and Remote Sensing}, vol.~61, pp. 1--16, 2023.

\bibitem{zhang2022prototype}
B.~Zhang, X.~Zhang, M.~Pun, and M.~Liu, ``Prototype-based clustered federated learning for semantic segmentation of aerial images,'' \emph{IEEE International Geoscience and Remote Sensing Symposium}, pp. 2227--2230, 2022.

\bibitem{buyuktacs2023federated}
B.~B{\"u}y{\"u}kta{\c{s}}, G.~Sumbul, and B.~Demir, ``Federated learning across decentralized and unshared archives for remote sensing image classification,'' \emph{IEEE Geoscience and Remote Sensing Magazine}, vol.~12, pp. 64--80, 2024.

\bibitem{buyuktacs2024transformer}
B.~B{\"u}y{\"u}kta{\c{s}}, K.~Weitzel, S.~V{\"o}lkers, F.~Zailskas, and B.~Demir, ``Transformer-based federated learning for multi-label remote sensing image classification,'' \emph{IEEE International Geoscience and Remote Sensing Symposium}, 2024.

\bibitem{li2024towards}
J.~Li, M.~Gong, Z.~Liu, S.~Wang, Y.~Zhang, Y.~Zhou, and Y.~Gao, ``Towards multi-party personalized collaborative learning in remote sensing,'' \emph{IEEE Transactions on Geoscience and Remote Sensing}, 2024.

\bibitem{wang2024personalized}
S.~Wang, J.~Li, Z.~Liu, M.~Gong, Y.~Zhang, Y.~Zhao, B.~Deng, and Y.~Zhou, ``Personalized multi-party few-shot learning for remote sensing scene classification,'' \emph{IEEE Transactions on Geoscience and Remote Sensing}, vol.~62, pp. 1--16, 2024.

\bibitem{chen2024free}
S.~Chen, T.~Shu, H.~Zhao, J.~Wang, S.~Ren, and L.~Yang, ``Free lunch for federated remote sensing target fine-grained classification: A parameter-efficient framework,'' \emph{Elsevier Knowledge-Based Systems}, 2024.

\bibitem{zhang2023local}
Z.~Zhang, X.~Ma, and J.~Ma, ``Local differential privacy based membership-privacy-preserving federated learning for deep-learning-driven remote sensing,'' \emph{Remote Sensing}, vol.~15, no.~20, p. 5050, 2023.

\bibitem{zhu2023privacy}
J.~Zhu, J.~Wu, A.~K. Bashir, Q.~Pan, and Y.~Wu, ``Privacy-preserving federated learning of remote sensing image classification with dishonest-majority,'' \emph{IEEE Journal of Selected Topics in Applied Earth Observations and Remote Sensing}, 2023.

\bibitem{qayyum2022collaborative}
A.~Qayyum, K.~Ahmad, M.~A. Ahsan, A.~Al-Fuqaha, and J.~Qadir, ``Collaborative federated learning for healthcare: Multi-modal covid-19 diagnosis at the edge,'' \emph{IEEE Open Journal of the Computer Society}, vol.~3, pp. 172--184, 2022.

\bibitem{che2023multimodal}
L.~Che, J.~Wang, Y.~Zhou, and F.~Ma, ``Multimodal federated learning: A survey,'' \emph{Sensors}, vol.~23, no.~15, p. 6986, 2023.

\bibitem{chen2024feddat}
H.~Chen, Y.~Zhang, D.~Krompass, J.~Gu, and V.~Tresp, ``Fed{DAT}: An approach for foundation model finetuning in multi-modal heterogeneous federated learning,'' in \emph{Proceedings of the AAAI Conference on Artificial Intelligence}, vol.~38, no.~10, 2024, pp. 11\,285--11\,293.

\bibitem{sun2023single}
X.~Sun, Y.~Tian, W.~Lu, P.~Wang, R.~Niu, H.~Yu, and K.~Fu, ``From single-to multi-modal remote sensing imagery interpretation: A survey and taxonomy,'' \emph{Science China Information Sciences}, vol.~66, no.~4, p. 140301, 2023.

\bibitem{huba2022papaya}
D.~Huba, J.~Nguyen, K.~Malik, R.~Zhu, M.~Rabbat, A.~Yousefpour, C.-J. Wu, H.~Zhan, P.~Ustinov, H.~Srinivas \emph{et~al.}, ``Papaya: Practical, private, and scalable federated learning,'' \emph{Proceedings of Machine Learning and Systems}, vol.~4, pp. 814--832, 2022.

\bibitem{chen2021communication}
M.~Chen, N.~Shlezinger, H.~V. Poor, Y.~C. Eldar, and S.~Cui, ``Communication-efficient federated learning,'' \emph{Proceedings of the National Academy of Sciences}, vol. 118, no.~17, 2021.

\bibitem{roy2019unsupervised}
S.~Roy, A.~Siarohin, E.~Sangineto, S.~R. Bulo, N.~Sebe, and E.~Ricci, ``Unsupervised domain adaptation using feature-whitening and consensus loss,'' \emph{IEEE Conference on Computer Vision and Pattern Recognition}, pp. 9471--9480, 2019.

\bibitem{xu2023personalized}
J.~Xu, X.~Tong, and S.-L. Huang, ``Personalized federated learning with feature alignment and classifier collaboration,'' \emph{International Conference on Learning Representations}, 2023.

\bibitem{quan2023novel}
D.~Quan, H.~Wei, S.~Wang, Y.~Gu, B.~Hou, and L.~Jiao, ``A novel coarse-to-fine deep learning registration framework for multi-modal remote sensing images,'' \emph{IEEE Transactions on Geoscience and Remote Sensing}, vol.~61, 2023.

\bibitem{sumbul2022novel}
G.~Sumbul, M.~M{\"u}ller, and B.~Demir, ``A novel self-supervised cross-modal image retrieval method in remote sensing,'' \emph{IEEE International Conference on Image Processing}, pp. 2426--2430, 2022.

\bibitem{chen2020simple}
T.~Chen, S.~Kornblith, M.~Norouzi, and G.~Hinton, ``A simple framework for contrastive learning of visual representations,'' in \emph{International Conference on Machine Learning}.\hskip 1em plus 0.5em minus 0.4em\relax PMLR, 2020, pp. 1597--1607.

\bibitem{FedAvg}
H.~B. McMahan, E.~Moore, D.~Ramage, S.~Hampson, and B.~A. Arcas, ``Communication-efficient learning of deep networks from decentralized data,'' \emph{International Conference on Artificial Intelligence and Statistics}, pp. 1273--1282, 2017.

\bibitem{li2021model}
Q.~Li, B.~He, and D.~Song, ``Model-contrastive federated learning,'' \emph{IEEE/CVF Conference on Computer Vision and Pattern Recognition}, pp. 10\,713--10\,722, 2021.

\end{thebibliography}
\end{document}